\documentclass[lettersize,journal]{IEEEtran}
\usepackage{amsmath,amsfonts}
\usepackage{algorithmic}
\usepackage{algorithm}
\usepackage{array}
\usepackage[caption=false,font=normalsize,labelfont=sf,textfont=sf]{subfig}
\usepackage{textcomp}
\usepackage{stfloats}
\usepackage{url}
\usepackage{verbatim}
\usepackage{graphicx}
\usepackage{float}
\usepackage{booktabs}
\usepackage{multirow}

\def\BibTeX{{\rm B\kern-.05em{\sc i\kern-.025em b}\kern-.08em
    T\kern-.1667em\lower.7ex\hbox{E}\kern-.125emX}}
\usepackage{balance}
\usepackage[colorlinks,bookmarksopen,bookmarksnumbered,citecolor=blue, linkcolor=blue, urlcolor=blue]{hyperref}
\usepackage{enumitem}
\makeatletter
\renewcommand{\maketag@@@}[1]{\hbox{\m@th\normalsize\normalfont#1}}%
\makeatother
\usepackage{amssymb}
\usepackage{nomencl}
\makenomenclature
\usepackage{etoolbox}
\usepackage{cuted}
\usepackage{soul, color, xcolor}

\begin{document}

\title{SE-BSFV: Online Subspace Learning based Shadow Enhancement and Background Suppression for ViSAR under Complex Background}
\author{Shangqu Yan, Chenyang Luo, Yaowen Fu, Wenpeng Zhang, Wei Yang, Ruofeng Yu
\thanks{The authors are with the College of Electronic Science and Technology, National University of Defense Technology, Changsha 410073, China. (e-mail: shangqu\_yan@163.com)}}
\maketitle

\begin{abstract}
Video synthetic aperture radar (ViSAR) has attracted substantial attention in the moving target detection (MTD) field due to its ability to continuously monitor changes in the target area. In ViSAR, the moving targets' shadows will not offset and defocus, which is widely used as a feature for MTD. However, the shadows are difficult to distinguish from the low scattering region in the background, which will cause more missing and false alarms. Therefore, it is worth investigating how to enhance the distinction between the shadows and background. In this study, we proposed the Shadow Enhancement and Background Suppression for ViSAR (SE-BSFV) algorithm. The SE-BSFV algorithm is based on the low-rank representation (LRR) theory and adopts online subspace learning technique to enhance shadows and suppress background for ViSAR images. Firstly, we use a registration algorithm to register the ViSAR images and utilize Gaussian mixture distribution (GMD) to model the ViSAR data. Secondly, the knowledge learned from the previous frames is leveraged to estimate the GMD parameters of the current frame, and the Expectation-maximization (EM) algorithm is used to estimate the subspace parameters. Then, the foreground matrix of the current frame can be obtained. Finally, the alternating direction method of multipliers (ADMM) is used to eliminate strong scattering objects in the foreground matrix to obtain the final results. The experimental results indicate that the SE-BSFV algorithm significantly enhances the shadows' saliency and greatly improves the detection performance while ensuring efficiency compared with several other advanced pre-processing algorithms.
\end{abstract}


\renewcommand\nomgroup[1]{%
  \item[\bfseries
  \ifstrequal{#1}{A}{Abbreviations}{%
  \ifstrequal{#1}{S}{Symbols}{}}%
]}
\mbox{}

\nomenclature[A, 01]{ViSAR}{Video Synthetic Aperture Radar}
\nomenclature[A, 02]{SNL}{Sandia National Laboratories}
\nomenclature[A, 03]{SAR}{Synthetic Aperture Radar}
\nomenclature[A, 04]{MTD}{Moving Target Detection}
\nomenclature[A, 05]{MF}{Median Filtering}
\nomenclature[A, 06]{SAR-BM3D}{SAR-block Matching 3 Dimensions}
\nomenclature[A, 07]{V-BM3D}{Video Block Matching 3 Dimensions}
\nomenclature[A, 08]{HESE}{Shadow Enhancement Algorithm Based on Histogram Equalization}
\nomenclature[A, 09]{LRSD}{Low-rank and Sparse Matrix Decomposition}
\nomenclature[A, 10]{SBN-3D-SD}{Shadow-background Noise 3D Spatial Decomposition}
\nomenclature[A, 11]{LRR}{Low-rank Representation}
\nomenclature[A, 12]{SE-BSFV}{Shadow Enhancement and Background Suppression for ViSAR}
\nomenclature[A, 13]{GMD}{Gaussian Mixture Distribution}
\nomenclature[A, 14]{ADMM}{Alternating Direction Method of Multipliers}
\nomenclature[A, 15]{GMD-LRR}{LRR model based on the GMD}
\nomenclature[A, 16]{EM}{Expectation-maximization}
\nomenclature[A, 17]{SURF}{Speeded-up Robust Features}
\nomenclature[A, 18]{SGD}{Stochastic Gradient Descent}
\nomenclature[A, 19]{NMS}{Non-maximum Suppression}
\nomenclature[A, 20]{AP}{Average Precision}
\nomenclature[A, 21]{PR}{Precision-Recall}
\nomenclature[A, 22]{IoU}{Intersection over Union}
\nomenclature[A, 23]{SNR}{Signal-to-noise Ratios}
\nomenclature[A, 24]{ALS}{Alternating Least Squares}
\nomenclature[A, 25]{CDF}{Cumulative Distribution Function}

\nomenclature[S, 01]{$B$}{The low rank matrix}
\nomenclature[S, 02]{$S$}{The sparse matrix}
\nomenclature[S, 03]{$N$}{The noise matrix}
\nomenclature[S, 04]{$X$}{The ViSAR data matrix}
\nomenclature[S, 05]{$W$}{The indicator matrix}
\nomenclature[S, 06]{$\odot$}{The Hadamard product}
\nomenclature[S, 07]{$U$}{The basis matrix}
\nomenclature[S, 08]{$V$}{The coefficient matrix}
\nomenclature[S, 09]{$K$}{The number of total components of the GMD}
\nomenclature[S, 10]{${{\pi }_{k}}$}{The weight of the $k-th$ Gaussian distribution}
\nomenclature[S, 11]{$\sigma _{k}^{2}$}{The variance of the $k-th$ Gaussian distribution}
\nomenclature[S, 12]{$\Omega$}{The set of all Gaussian components' weights}
\nomenclature[S, 13]{$\Lambda$}{The set of all Gaussian components' variances}
\nomenclature[S, 14]{$\Theta $}{The index set of non-missing data of matrix $X$}
\nomenclature[S, 15]{$\eta $}{The regularization parameter}

\printnomenclature

\section{Introduction}
\IEEEPARstart{T}{he} concept of ViSAR was proposed by SNL in 2003, and high frame rate imaging is achieved on an airborne platform \cite{r1,r2}. The emergence of ViSAR provides a pivotal technology for high-resolution video surveillance of the target area. ViSAR not only has the characteristics of traditional SAR such as long-range detection and strong environmental adaptability, but also offers the advantage of continuously observing the target area. As an extension of the SAR system, ViSAR holds significant application potential for dynamic monitoring \cite{r3,r4,r5,r6,z7}. However, during the ViSAR imaging, the moving target usually appears offset and defocused in the azimuth direction due to Doppler modulation \cite{10551832}, while its shadow appears in the real position and maintains a constant gray level throughout the whole synthetic aperture time \cite{r7}. Consequently, detecting the moving targets' shadows is an effective strategy to enhance the performance of MTD in ViSAR \cite{r8,r9,r10}.

However, it is difficult to distinguish the moving targets' shadows from the low scattering region of complex background in ViSAR images. Therefore, detecting the moving target's shadow can be classified as a “Weak Target” detection problem. The application of pre-processing techniques for ViSAR images in MTD can mitigate issues related to missing and false alarms caused by “Weak Targets” to some extent \cite{r11,r12,r13}.

In the research of pre-processing for ViSAR images, two categories emerge: \textbf{de-speckling} and \textbf{moving targets' shadows enhancement}. The purpose of de-speckling is to effectively suppress speckle noise in ViSAR images while preserving the details of moving targets' shadows. The purpose of moving targets' shadows enhancement is to enhance the representation of their characteristics while suppressing the low scattering region in the background. From various perspectives, both categories ultimately aim to better retain the characteristics of moving targets' shadows, improve the moving targets' shadows' quality, and provide a guarantee for subsequent target detection applications.

Currently, there are primarily three pre-processing algorithms for de-speckling ViSAR images: MF \cite{r14}, SAR-BM3D \cite{r15}, and V-BM3D algorithms \cite{r16}. These algorithms can filter speckle noise to a certain extent, but they can't fully integrate the characteristics of ViSAR data to improve their effectiveness. When the noise standard deviation is too large, the performance of the above algorithms is seriously reduced. To enhance the contrast between moving targets' shadows and their surrounding environment, some researchers have used the HESE algorithm \cite{r13}. Although the HESE algorithm effectively improves the contrast of moving targets' shadows, this global gray averaging operation will enhance stationary objects' shadows and noise.

Benefiting from the results of LRSD in computer vision, image processing, pattern recognition, and other fields, some researchers have begun applying the LRSD to enhance the moving targets' shadows in ViSAR. The LRSD algorithm is first applied in reference \cite{r17} to separate the background and foreground for ViSAR. However, this study can't model the noise in the ViSAR images. As a result, the moving targets' shadows, stationary targets' shadows, and noise are grouped into the foreground image. References \cite{r18} and \cite{r19} proposed the SBN-3D-SD algorithm to improve the contrast between the moving targets' shadows and background. The algorithm uses the sparsity characteristic of moving targets' shadows, the low-rank characteristic of the background, and the Gaussian characteristic of noise to decompose the three-dimensional space so that the foreground image is less affected by the stationary targets' shadows and noise. In reference \cite{r20}, the sparse constraint is replaced by the total variation constraint, and the dynamic background constraint term is introduced along with the correlation suppression term. This modification results in a purer foreground image. Although the above pre-processing algorithms are effective, there are practical problems such as large computational complexity and inaccurate estimation of low-rank matrix.

To address the shortcomings in the above algorithms, we have introduced online subspace learning technique into the LRR model and proposed a pre-processing algorithm: SE-BSFV. Experimental results show that the SE-BSFV algorithm can enhance the saliency of moving targets' shadows and suppress background while ensuring efficiency. Firstly, the SE-BSFV algorithm registers ViSAR images and forms a data matrix, and the pixel distribution of each ViSAR image is modeled as the GMD. Secondly, the subspace parameters and GMD parameters of the current frame are updated by using the knowledge of the previous frames, and the foreground matrix is output by iterative updating frame by frame. Finally, considering that the obtained foreground matrix still contains some strong scattering objects, the ADMM algorithm is used to remove them, and the ViSAR images after shadow enhancement and background suppression are obtained.

\section{Previous Work}
\subsection{The SBN-3D-SD Algorithm}
 The SBN-3D-SD algorithm \cite{r18,r19} is a variant of the LRSD algorithm, which can be formulated as a problem of minimizing the low-rank, ${{L}_{0}}$ norm and Frobenius norm. Generally, the nuclear norm and ${{L}_{1}}$ norm are used to replace the rank function and ${{L}_{0}}$ norm. The problem could be formulated as:
\begin{equation}
\tag*{(1)}
\label{eq1}
 \underset{B,S,N}{\mathop{\min }}\,{{\left\| B \right\|}_{*}}+\xi {{\left\| S \right\|}_{{{L}_{1}}}}+\gamma \left\| N \right\|_{F}^{2} \, s.t.X=B+S+N  \\
\end{equation}
where ${{\left\| \cdot \right\|}_{*}}$ denotes the nuclear norm, ${{\left\| \cdot  \right\|}_{{{L}_{1}}}}$ denotes the ${{L}_{1}}$ norm, and ${{\left\| \cdot  \right\|}_{F}}$ denotes the Frobenius norm. The $N,S,B,X\in {{\Re }^{d\times n}}$, $d$ represents the total number of pixels in a ViSAR image, and $n$ represents the number of ViSAR images to be processed. 

Then the ADMM algorithm \cite{r21} is used to solve \ref{eq1}, and the foreground image with moving targets' shadows is obtained. However, since the nuclear norm is the sum of all singular values, its results may deviate significantly from the true value, which will cause the low-rank matrix to be inaccurate. Simultaneously, the computation and memory requirements are amplified due to the extensive matrix dimensions resulting from the complete ViSAR sequence calculation.

\subsection{The GMD-LRR Algorithm}
To solve the problem caused by the nuclear norm, reference \cite{r22} adopts the GMD-LRR algorithm to make the estimated low-rank matrix more accurate. In comparison with the LRSD and SBN-3D-SD algorithms, the GMD-LRR algorithm does not need penalty parameters and can also avoid the bias problem caused by the nuclear norm. In the context of this study, the flow of the GMD-LRR algorithm can be described as follows.

Let ${X=[{{x}_{1}},\cdots ,{{x}_{i}},\cdots ,{{x}_{n}}]\in {{\Re }^{d\times n}}}$ be a given ViSAR data matrix, an LRR problem can be formulated as follows \cite{r22}:
\begin{equation}
\tag*{(2)}
\label{eq2}
\underset{U,V}{\mathop{\min }}\,{{\left\| W\odot (X-U{{V}^{T}}) \right\|}_{{{L}_{p}}}} \\
\end{equation}
where $U\in {{\Re }^{d\times r}}$, $V\in {{\Re }^{n\times r}}$, $r\le \min (d,n)$ represents the low-rank property of $U{{V}^{T}}$, and ${{\left\| \cdot  \right\|}_{{{L}_{p}}}}$ represents the ${{L}_{p}}$ norm. From the generative perspective, each element ${{x}_{ij}}$ of the data matrix $X$ can be modeled as:
\begin{equation}
\tag*{(3)}
\label{eq3}
{{x}_{ij}}={{u}_{i}}{{({{v}_{j}})}^{T}}+{{\varepsilon }_{ij}} \\
\end{equation}
where ${{u}_{i}}$ and ${{v}_{j}}$ are the $i-th$ and $j-th$ row vectors of $U$ and $V$, respectively, and ${{\varepsilon }_{ij}}$ is the residual term in ${{x}_{ij}}$.

To enhance the model's robustness against complex backgrounds, the residual term ${{\varepsilon }_{ij}}$ can be represented as a parametric probability distribution, and the GMD is utilized for modeling the residual term:
\begin{equation}
\tag*{(4)}
\label{eq4}
{{\varepsilon }_{ij}}\sim \sum\nolimits_{k=1}^{K}{{{\pi }_{k}}P({{\varepsilon }_{ij}}\left| 0,\sigma _{k}^{2} \right.)} \\
\end{equation}
where ${P({{\varepsilon }_{ij}}\left| 0,\sigma _{k}^{2} \right.)=\frac{1}{\sqrt{2\pi }{{\sigma }_{k}}}\exp (-\frac{\varepsilon _{ij}^{2}}{2\sigma _{k}^{2}})]}$ is the distribution of the $k-th$ Gaussian component. Therefore, the probability of each element ${{x}_{ij}}$ of the matrix $X$ can be written as follows:
\begin{equation}
\tag*{(5)}
\label{eq5}
p({{x}_{ij}}\left| {{u}_{i}} \right.,{{v}_{j}},\Omega ,\Lambda )=\sum\nolimits_{k=1}^{K}{{{\pi }_{k}}}P({{x}_{ij}}\left| {{u}_{i}}{{({{v}_{j}})}^{T}},\sigma _{k}^{2} \right.) \\
\end{equation}
where $\Omega =\left\{ {{\pi }_{k}} \right\}_{k=1}^{K}$, and $\Lambda =\left\{ \sigma _{k}^{2} \right\}_{k=1}^{K}$. The likelihood of the matrix $X$ can be written as follows:
\begin{equation}
\tag*{(6)}
\label{eq6}
\begin{aligned}
& p(X\left| U \right.,V,\Omega ,\Lambda )\\
& =\prod\limits_{i,j\in \Theta }{\sum\nolimits_{k=1}^{K}{{{\pi }_{k}}P[{{x}_{ij}}{{\left| {{u}_{i}}({{v}_{j}}) \right.}^{T}},\sigma _{k}^{2}]}} \\ 
 & =\prod\limits_{i,j\in \Theta }{p[{{x}_{ij}}{{\left| {{u}_{i}}({{v}_{j}}) \right.}^{T}},\Omega ,\Lambda ]}  \\
 \end{aligned}
\end{equation}
where $i$ and $j$ are the indices of row and column in the set $\Theta $. The GMD-LRR algorithm's goal is to maximize the log-likelihood function for the GMD parameters $\Omega $, $\Lambda $ and the subspace parameters $U$, $V$:
\begin{equation}
\tag*{(7)}
\label{eq7}
\begin{aligned}
  & \underset{U,V,\Omega ,\Lambda }{\mathop{\max }}\,\mathcal{L}(U,V,\Omega ,\Lambda )= \\ 
 & \sum\nolimits_{i,j\in \Theta }{\ln \sum\nolimits_{k=1}^{K}{{{\pi }_{k}}P[{{x}_{ij}}{{\left| {{u}_{i}}({{v}_{j}}) \right.}^{T}},\sigma _{k}^{2}]}} \\ 
 \end{aligned}
\end{equation}

The EM algorithm \cite{r23} can be used to find the maximum likelihood parameter $(U,V,\Omega ,\Lambda )$. The derivation of the EM algorithm is in \ref{Appendix-A}. After solving all the parameters, \ref{eq2} can be naturally solved. In this study, we use the ${{L}_{2}}$ norm instead of the ${{L}_{p}}$ norm to solve \ref{eq2}. 

\section{Methodology}
\begin{figure*}[ht]
\centering
\includegraphics[width=1\textwidth]{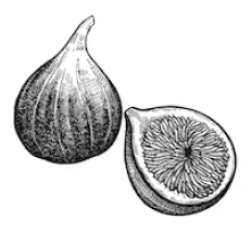}
\caption{The flowchart of the SE-BSFV algorithm.}
\label{figure1}
\end{figure*}
Fig. \ref{figure1} shows the overall flowchart of the SE-BSFV algorithm, which can be divided into three parts: registration based on the SURF algorithm, online subspace learning, and the ADMM processing. In this section, we will present these three parts of the SE-BSFV algorithm in detail.

\subsection{Registration Based on the SURF Algorithm}
Since circular SAR imaging is used in the SNL dataset, the viewing angle in the scene inevitably changes. According to our calculations, the rotation angle of the whole dataset ranges from 0 to 127 degrees. To maintain the assumption of the low-rank characteristics of the background, we used the SURF algorithm \cite{r24} to register ViSAR images (performed every 100 frames, where the first frame of each subvideo is the reference image).

\subsection{Online Subspace Learning}
The analysis in \ref{Appendix-B} demonstrates that the background of ViSAR data exhibits low-rank characteristics and can be effectively modeled using a GMD. Therefore, we can employ the LRR model to calculate the low-rank matrix of ViSAR data. To reduce calculation and memory consumption, we utilize the concept of online learning \cite{r25} to adjust the subspace parameters of the GMD-LRR algorithm, so that it can process the ViSAR data of each image in real-time. For instance, under the regularization of the foreground or background knowledge learned before the frame $t-1$, the specific foreground GMD distribution (GMD parameters) and specific background subspace distribution (subspace parameters) are inferred for the frame $t$. 

Specifically: the parameters ${{\Omega }_{t}}=\left\{ {{\pi }_{k,t}} \right\}_{k=1}^{K}$ and ${{\Lambda }_{t}}=\left\{ \sigma _{k,t}^{2} \right\}_{k=1}^{K}$ of the frame $t$ can be obtained from the parameters ${{\Omega }_{t-1}}$, ${{\Lambda }_{t-1}}$, and ${{N}_{k,t-1}}=\sum\nolimits_{i,j}{{{\gamma }_{ijk,t-1}}} $. Where the parameter ${{N}_{k,t-1}}=\sum\nolimits_{i,j}{{{\gamma }_{ijk,t-1}}}$ has the same meaning as in \ref{eqA.3} in \ref{Appendix-A}. The ${{V}_{t}}$ and ${{U}_{t}}$ of the frame $t$ are obtained from the subspace ${{U}_{t-1}}$. The solution model for the parameters $\Omega $ and $\Lambda $ are as follows:
\begin{equation}
\tag*{(8)}
\label{eq8}
\left\{ {{\Omega }_{t}},{{\Lambda }_{t}} \right\}=\arg \underset{\Omega ,\Lambda }{\mathop{\max }}\,\sum\nolimits_{j=1}^{t}{\ln p({{x}_{j}},{{z}_{j}}\left| {{U}_{j}},{{V}_{j}},\Omega ,\Lambda  \right.)}
\end{equation}
where ${{x}_{j}}$ is the d-dimensional column vector of the frame $j$, ${{z}_{j}}$ is the hidden variable of the frame $j$ (The calculation of ${{z}_{j}}$ can be derived from \ref{eqA.1} in \ref{Appendix-A}), ${{U}_{j}}$ and ${{V}_{j}}$ are the subspace and subspace coefficient vector of the frame $j$, respectively. The solution model for the parameter ${U}$ is as follows:
\begin{equation}
\tag*{(9)}
\label{eq9}
{{U}_{t}}=\arg \underset{U}{\mathop{\max }}\,\sum\nolimits_{j=1}^{t}{\ln p({{x}_{j}},{{z}_{j}}\left| U,{{V}_{j}},{{\Omega }_{j}},{{\Lambda }_{j}} \right.)}
\end{equation}

\ref{eq8} and \ref{eq9} can be written as the sum of the pixel knowledge distributions before the frame $t-1$ and at the frame $t$, respectively:
\begin{equation}
\tag*{(10)}
\label{eq10}
\begin{aligned}
& \sum\nolimits_{j=1}^{t}{\ln p({{x}_{j}},{{z}_{j}}\left| {{U}_{j}},{{V}_{j}},\Omega ,\Lambda  \right.)} \\
& =\sum\nolimits_{j=1}^{t-1}{\ln p({{x}_{j}},{{z}_{j}}\left| {{U}_{j}},{{V}_{j}},\Omega ,\Lambda  \right.)}\\
& +\ln p({{x}_{t}},{{z}_{t}}\left| {{U}_{t}},{{V}_{t}},\Omega ,\Lambda  \right.) \\ 
\end{aligned}
\end{equation}
\begin{equation}
\tag*{(11)}
\label{eq11}
\begin{aligned}
 & \sum\nolimits_{j=1}^{t}{\ln p({{x}_{j}},{{z}_{j}}\left| U,{{V}_{j}},{{\Omega }_{j}},{{\Lambda }_{j}} \right.)}\\
 & =\sum\nolimits_{j=1}^{t-1}{\ln p({{x}_{j}},{{z}_{j}}\left| U,{{V}_{j}},{{\Omega }_{j}},{{\Lambda }_{j}} \right.)}\\
 & +\ln p({{x}_{t}},{{z}_{t}}\left| U,{{V}_{t}},{{\Omega }_{t}},{{\Lambda }_{t}} \right.) \\ 
\end{aligned}
\end{equation}

According to the above model framework, the parameters $U,V,\Omega ,\Lambda $ are solved by using the knowledge of the previous frames to achieve online learning. The first term in \ref{eq10} represents the distribution of pixels before the frame $t-1$ and the second term represents the distribution of pixels in the frame $t$. The distribution of pixels before the frame $t-1$ can be expanded as follows (The derivation is based on \ref{eqA.2} in \ref{Appendix-A}):

\begin{strip}
\begin{equation}
\tag*{(12)}
\label{eq12}
\begin{aligned}
  & \sum\nolimits_{j=1}^{t-1}{\ln p({{x}_{j}},{{z}_{j}}\left| {{U}_{j}},{{V}_{j}},\Omega ,\Lambda  \right.)}  \\
  & =\sum\nolimits_{j=1}^{t-1} \Big\{ \sum\nolimits_{i=1}^{d}{\sum\nolimits_{k=1}^{K}{{{z}_{k,j}}}\ln {{\pi }_{k}} -\sum\nolimits_{k=1}^{K}{\sum\nolimits_{i=1}^{d}{{{z}_{k,j}}\ln {{\sigma }_{k}}}}}  -\frac{\sum\nolimits_{k=1}^{K}{\sum\nolimits_{i=1}^{d}{{{z}_{k,j}}{{[{{x}_{ij}}-{{u}_{i}}{{({{v}_{j}})}^{T}}]}^{2}}}}}{2\pi \sigma _{k}^{2}} \Big\} \\ 
 & =\sum\nolimits_{k=1}^{K}{\Big\{ {{N}_{k,t-1}}\ln {{\pi }_{k}} }-{{N}_{k,t-1}}\ln {{\sigma }_{k}} -\frac{{{N}_{k,t-1}}\frac{1}{{{N}_{k,t-1}}}\sum\nolimits_{j=1}^{t-1}{\sum\nolimits_{i=1}^{d}{{{z}_{k,j}}{{[{{x}_{ij}}-{{u}_{i}}{{({{v}_{j}})}^{T}}]}^{2}}}}}{2\pi \sigma _{k}^{2}} \Big\} \\ 
 & ={{N}_{t-1}}\sum\nolimits_{k=1}^{K}{{{\pi }_{k,t-1}}}\ln {{\pi }_{k}} -\sum\nolimits_{k=1}^{K}{{{N}_{k,t-1}}}(\ln {{\sigma }_{k}}+\frac{\sigma _{k,t-1}^{2}}{2\sigma _{k}^{2}}) \\ 
\end{aligned}
\end{equation}
\end{strip}
\noindent where ${{N}_{k,t-1}}=\sum\nolimits_{j=1}^{t-1}{\sum\nolimits_{i=1}^{d}{{{z}_{k,j}}}}$, ${{N}_{t-1}}=\sum\nolimits_{k=1}^{K}{{{N}_{k,t-1}}}$, $\sigma _{k,t-1}^{2}=\frac{\sum\nolimits_{j-1}^{t-1}{\sum\nolimits_{i-1}^{d}{{{z}_{k,j}}{{\left[ {{x}_{ij}}-{{u}_{i}}{{\left( {{v}_{j}} \right)}^{T}} \right]}^{2}}}}}{{{N}_{k,t-1}}}$, ${{\pi }_{k,t-1}}=\frac{{{N}_{k,t-1}}}{{{N}_{t-1}}}$, and ${{z}_{k,t}}$ represents the hidden variable ${{z}_{k}}$ for the frame $j$ within frames $1\sim t-1$.

According to \ref{eq8} and \ref{eq12}, we know that the parameters ${{\Omega }_{t}}$ and ${{\Lambda }_{t}}$ are affected by the parameters $\left\{ {{\pi }_{k,t-1}} \right\}_{k=1}^{K}$, $\left\{ \sigma _{k,t-1}^{2} \right\}_{k=1}^{K}$ and $\left\{ {{N}_{k,t-1}} \right\}_{k=1}^{K}$. Therefore, the parameters ${{\Omega }_{t-1}}$ and ${{\Lambda }_{t-1}}$ can be used to correct the parameters ${{\Omega }_{t}}$ and ${{\Lambda }_{t}}$ of the frame $t$. 

Using \ref{eq12} as the correction factor of the frame $t$, we can set ${{M}_{t}}(\Omega ,\Lambda )={{N}_{t-1}}\sum\nolimits_{k=1}^{K}{{{\pi }_{k,t-1}}}\ln {{\pi }_{k}}-\sum\nolimits_{k=1}^{K}{{{N}_{k,t-1}}}(\ln {{\sigma }_{k}}+\frac{\sigma _{k,t-1}^{2}}{2\sigma _{k}^{2}})$, and then \ref{eq10} can be rewritten as follows:
\begin{equation}
\tag*{(13)}
\label{eq13}
\begin{aligned}
  & \sum\nolimits_{j=1}^{t}{\ln p({{x}_{j}},{{z}_{j}}\left| {{U}_{j}},{{V}_{j}},\Omega ,\Lambda  \right.)} \\ 
 & =\ln p({{x}_{t}},{{z}_{t}}\left| {{U}_{t}},{{V}_{t}},\Omega ,\Lambda  \right.)+{{M}_{t}}(\Omega ,\Lambda ) \\ 
 & ={{L}_{t}}(\Omega ,\Lambda ) \\ 
\end{aligned}
\end{equation}

In \ref{eq13}, the first term is the likelihood term, which is the parameter of the frame $t$, and the second term is the regularization term of the GMD, which is the knowledge accumulation term before the frame $t-1$. By solving the partial derivative of the pixel knowledge from frame 1 to $t$, i.e., the partial derivative of ${{L}_{t}}(\Omega ,\Lambda )$, the pixel relationship between frame $t$ and before frame $t-1$ can be obtained, and then the frame $t$ can be predicted by using the knowledge before frame $t-1$:
\begin{equation}
\tag*{(14)}
\label{eq14}
\begin{aligned}
  & {{L}_{t}}(\Omega ,\Lambda )=\ln p({{x}_{t}},{{z}_{t}}\left| {{U}_{t}},{{V}_{t}},\Omega ,\Lambda  \right.)+{{M}_{t}}(\Omega ,\Lambda ) \\ 
 & =\sum\nolimits_{i=1}^{d}{\sum\nolimits_{k=1}^{K}{{{z}_{k,j}}\left[ \ln {{\pi }_{k}}-\frac{{{({{x}_{i,t}}-{{u}_{i}}{{v}^{T}})}^{2}}}{2\sigma _{k}^{2}}-\ln {{\sigma }_{k}} \right]}} \\ 
 & +{{N}_{t-1}}\sum\nolimits_{k=1}^{K}{{{\pi }_{k,t-1}}}\ln {{\pi }_{k}} \\ 
 & -\sum\nolimits_{k=1}^{K}{{{N}_{k,t-1}}}(\ln {{\sigma }_{k}}+\frac{\sigma _{k,t-1}^{2}}{2\sigma _{k}^{2}})
\end{aligned}
\end{equation}

The maximum value of ${{\sigma }_{k}}$ is solved as follows:
\begin{equation}
\tag*{(15)}
\label{eq15}
\begin{aligned}
& \frac{\partial {{L}_{t}}(\Omega ,\Lambda )}{\partial {{\sigma }_{k}}}=\sum\nolimits_{i=1}^{d}{{{z}_{k,j}}}[-\frac{1}{{{\sigma }_{k}}}+\frac{{{({{x}_{i,t}}-{{u}_{i}}{{v}^{T}})}^{2}}}{\sigma _{k}^{3}} \\ 
& +{{N}_{k,t-1}}(-\frac{1}{{{\sigma }_{k}}}+\frac{\sigma _{k,t-1}^{2}}{\sigma _{k}^{3}})]  =0
\end{aligned}
\end{equation}

From \ref{eq15}, the following can be obtained:
\begin{equation}
\tag*{(16)}
\label{eq16}
\begin{aligned}
\sigma _{k}^{2}=\frac{{{N}_{k,t-1}}\sigma _{k,t-1}^{2}+\sum\nolimits_{i=1}^{d}{{{z}_{k,j}}}{{({{x}_{i,t}}-{{u}_{i}}{{v}^{T}})}^{2}}}{{{N}_{k,t-1}}+\sum\nolimits_{i=1}^{d}{{{z}_{k,j}}}}
\end{aligned}
\end{equation}

Let $\sum\nolimits_{k=1}^{K}{{{\pi }_{k}}}=1$, and take the derivative of the Lagrangian constraint function \cite{r26} of ${{\pi }_{k}}$, the following can be obtained:
\begin{equation}
\tag*{(17)}
\label{eq17}
\begin{aligned}
& \frac{\partial [{{L}_{t}}(\Omega ,\Lambda )-\lambda (\sum\nolimits_{k=1}^{K}{{{\pi }_{k}}-1})]}{\partial {{\pi }_{k}}} \\ 
 & =\sum\nolimits_{i=1}^{d}{{{z}_{k,j}}\frac{1}{{{\pi }_{k}}}}+{{N}_{k,t-1}}\frac{1}{{{\pi }_{k}}}-\lambda  \\ 
 & =0 
\end{aligned}
\end{equation}

\ref{eq17} can be further derived as follows:
\begin{equation}
\tag*{(18)}
\label{eq18}
\sum\nolimits_{i=1}^{d}{{{z}_{k,j}}}+{{N}_{k,t-1}}-\lambda {{\pi }_{k}}=0
\end{equation}

According to \ref{eqA.3} and \ref{eq18}, we can deduce the values of ${{\pi }_{k}}$ and $\lambda $ of the frame $t$:
\begin{equation}
\tag*{(19)}
\label{eq19}
\left\{ \begin{array}{*{35}{l}}
   {{\pi }_{k}}=\frac{{{N}_{k,t-1}}+\sum\nolimits_{i=1}^{d}{{{z}_{k,j}}}}{\sum\nolimits_{k=1}^{K}{\left[ {{N}_{k,t-1}}+\left. \sum\nolimits_{i=1}^{d}{{{z}_{k,j}}} \right] \right.}}  \\
   \lambda =\sum\nolimits_{k=1}^{K}{\left[ {{N}_{k,t-1}}+\sum\nolimits_{i=1}^{d}{{{z}_{k,j}}} \right]}  \\
\end{array} \right.\
\end{equation}

Through the above derivation, we can update the parameters $\Omega $ and $\Lambda $ in an online learning way. Once we have the $\Omega $ and $\Lambda $ for the current frame, we can update the parameter $U$. According to \ref{eqA.4}, the second term of \ref{eq11} can be rewritten as follows:
\begin{equation}
\tag*{(20)}
\label{eq20}
\begin{aligned}
& \ln p({{x}_{t}},{{z}_{t}}\left| U,{{V}_{t}},{{\Omega }_{t}},{{\Lambda }_{t}} \right.) \\ 
 & =\sum\nolimits_{i=1}^{d}{\sum\nolimits_{k=1}^{K}{{{z}_{k,t}}}}\left\{ -\frac{{{\left[ {{x}_{i,t}}-{{u}_{i}}{{({{v}_{t}})}^{T}} \right]}^{2}}}{2\sigma _{k,t}^{2}} \right\} \\ 
 & =-\sum\nolimits_{i=1}^{d}{\left( \sum\nolimits_{k=1}^{K}{\frac{{{z}_{k,t}}}{2\sigma _{k,t}^{2}}} \right)}{{\left[ {{x}_{i,t}}-{{u}_{i}}{{({{v}_{t}})}^{T}} \right]}^{2}} \\ 
 & =-{{\left\| {{W}_{t}}\odot \left[ {{X}_{t}}-U{{({{V}_{t}})}^{T}} \right] \right\|}_{{{L}_{2}}}}
\end{aligned}
\end{equation}
where ${{\sigma }_{k,t}}$ is the standard deviation ${{\sigma }_{k}}$ of the $k-th$ Gaussian component of the frame $t$, ${{z}_{k,t}}$ is ${{z}_{k}}$ of the frame $t$, ${{W}_{t}}$ is the index matrix of the frame $t$, and ${{X}_{t}}$ is the data of the frame $t$.

According to \ref{eqA.6}, finding the minimum of \ref{eq20} is a weighted least squares problem whose closed-form solution is as follows \cite{r25}:
\begin{equation}
\tag*{ (21)}
\label{eq21}
{{v}_{t}}={{[{{U}^{T}}diag{{({{W}_{t}})}^{2}}U]}^{-1}}{{U}^{T}}diag{{({{W}_{t}})}^{2}}{{x}_{t}}
\end{equation}

Although ${{u}_{i,t}}$ is the subspace component of frame $t$, it follows from \ref{eq11} that it is essentially the cumulative result from frame 1 to $t$. Therefore, ${{u}_{i,t}}$ can be written as follows \cite{r25}:
\begin{equation}
\tag*{(22)}
\label{eq22}
\begin{aligned}
& {{u}_{i,t}}\\
&={{[\sum\nolimits_{j=1}^{t}{w_{i,j}^{2}{{v}_{j}}{{({{v}_{j}})}^{T}}}]}^{-1}}[\sum\nolimits_{j=1}^{t}{w_{i,j}^{2}{{x}_{i,j}}{{({{v}_{j}})}^{T}}}]  \\
&={{[A_{i,t-1}^{-1}+w_{i,t}^{2}{{v}_{t}}{{({{v}_{t}})}^{T}}]}^{-1}}[A_{i,t-1}^{-1}{{u}_{i,t-1}}+w_{i,t}^{2}{{x}_{i,t}}{{({{v}_{t}})}^{T}}]
\end{aligned}
\end{equation}
\noindent where $A_{i,t-1}^{-1}=\sum\nolimits_{j=1}^{t-1}{w_{i,j}^{2}{{v}_{j}}}{{({{v}_{j}})}^{T}}$, ${{x}_{i,j}}$ denotes the $i-th$ pixel in the ${{x}_{j}}$ vector, the definition of $w_{i,j}$ is given in \ref{eqA.5}.

Let $A_{i,t}^{-1}=A_{i,t-1}^{-1}+w_{i,t}^{2}{{v}_{t}}{{({{v}_{t}})}^{T}}$, ${{B}_{i,t}}=A_{i,t-1}^{-1}{{u}_{i,t-1}}+w_{i,t}^{2}{{x}_{i,t}}{{({{v}_{t}})}^{T}}$, and then ${{u}_{i,t}}={{A}_{i,t}}{{B}_{i,t}}$. Therefore, the expressions for ${{A}_{i,t}}$ and ${{B}_{i,t}}$ are as follows:
\begin{equation}
\tag*{(23)}
\label{eq23}
\left\{ \begin{array}{*{35}{l}}
   {{A}_{i,t}}=\frac{{{A}_{i,t-1}}}{1+w_{i,t-1}^{2}{{v}_{t}}{{({{v}_{t}})}^{T}}{{A}_{i,t-1}}}  \\
   {{B}_{i,t}}={{B}_{i,t-1}}+w_{i,t}^{2}{{x}_{i,t}}{{({{v}_{t}})}^{T}}  \\
\end{array} \right.
\end{equation}
 where ${{B}_{i,t-1}}={{\sum\nolimits_{j=1}^{t-1}{w_{i,j}^{2}{{x}_{i,j}}({{v}_{j}})}}^{T}}$.
 
For the subspace ${{u}_{i,t}}$, only needs to compute ${{A}_{i,t}}$ and ${{B}_{i,t}}$. The above update method can avoid the matrix inversion calculation and ensure the efficiency of the algorithm. After obtaining ${{u}_{i,t}}$, the sparse pixel distribution (foreground matrix) ${{x}_{i,t}}-{{u}_{i,t}}{{v}_{t}}^{T}$ can be calculated. 

 \subsection{The ADMM Processing}
Following the above processing, the ideal situation is that the background and foreground are completely separated. However, certain strongly scattering objects may always be in the foreground matrix. Fortunately, the moving targets' shadows have a strong linear correlation in the data matrix composed of all foreground matrices, which can be projected onto a low-dimensional subspace. Therefore, to filter out the strong scattering objects, the ADMM algorithm \cite{r21} is used to encode the smoothness of all foreground matrixes. The final expression is as follows:
\begin{equation}
\tag*{(24)}
\label{eq24}
\begin{aligned}
  & \underset{S,O}{\mathop{\min }}\,{{\left\| S \right\|}_{*}}+\eta {{\left\| O \right\|}_{{{L}_{1}}}} \\
 &  s.t. \left\{ {{x}_{i,t}}-{{u}_{i,t}}{{v}_{t}}^{T} \right\}_{t=1}^{n}=S+O 
\end{aligned}
\end{equation}
where $O$ denotes the strong scattering objects, and $S$ denotes the moving targets' shadows.

A summary of the SE-BSFV algorithm is shown in Algorithm 1 below.
\begin{algorithm}[H] 
\caption{The SE-BSFV algorithm.} \label{alg1}
\begin{algorithmic}
\STATE
\REQUIRE
\STATE {the GMD parameters:${\Omega}_{t-1}, {\Lambda}_{t-1}, {N}_{t-1},{K}$};\\
\STATE {the model variabels:$\{{A}_{i, t-1}\}^{d}_{i=1}, \{{B}_{i, t-1}\}^{d}_{i=1}$};\\
\STATE {the subspace:${U}_{t-1}$};\\
\STATE {the data:${x}_{t}$};\\
\STATE	{the ${x}_{t}$ has been registered by the SURF algorithm};\\
\STATE {the nonnegative regularization parameter:${\eta}$}.
\ENSURE {${S}$}\\ 
\textbf{initialize:} Set $\{{\Omega}, {\Lambda}\} = \{{\Omega}_{t-1}, {\Lambda}_{t-1}\}$\\
\WHILE{{not converged}}
\STATE  {E-step: update ${r}_{itk}$ via \ref{eqA.1}},\\
\STATE  {M-step: update $\{{\Omega}_{t}, {\Lambda}_{t}, {N}_{t}\}$ via \ref{eq16} and \ref{eq19}},\\
\STATE  {update ${\nu}_{t}$ via \ref{eq21}}.\\
\ENDWHILE
\FOR{each ${{u}_{i,t}|}^{d}_{i=1}$}
\STATE {update $\{{A}_{i, t}\}^{d}_{i=1}, \{{B}_{i, t}\}^{d}_{i=1}$ via \ref{eq23}},\\
\STATE {update ${u}_{i,t}$ via ${u}_{i,t}=\{{A}_{i, t}\}^{d}_{i=1} \{{B}_{i, t}\}^{d}_{i=1}$}.\\
\ENDFOR
\STATE {compute $\mathop{min}\limits_{S,O}{||S||}_{*}+{\eta{||O||}_{L1}}$,\\
\STATE {s.t. $\{{x}_{i,t}-{u}_{i,t}{v}^{T}_{t}\}^{n}_{t=1}={S}+{O}$} via ADMM algorithm.}\\      
\end{algorithmic}
\end{algorithm}

\section{Experiments}
To evaluate the performance of the SE-BSFV algorithm, we compare it with the MF, SAR-BM3D, V-BM3D, HESE, LRSD, and SBN-3D-SD algorithms. Firstly, we performed shadow enhancement and background suppression experiments using the SNL's ViSAR data. Secondly, to validate the effectiveness of the proposed algorithm for MTD, we used the traditional detection algorithms and deep learning-based detectors for detection verification, respectively.

\subsection{Data and Experimental Equipment}
All experiments are conducted using the ViSAR dataset published by SNL \cite{r27}, which consists of 900 ViSAR images with 720*660 resolution for each image. The data of the ground truths of moving targets' shadows are annotated by us using LabelMe software. The shadow enhancement experiment and the detection experiment with traditional detection algorithm are implemented in MATLAB R2023a and run on a PC with an Intel i9-11900H processor and 32GB of RAM. The detection experiments using deep learning-based detectors are implemented in PyCharm using PyTorch (CUDA-11.2 and CuDNN-8.0.5) and run on a server with two NVIDIA GeForce RTX 3090 GPUs.

\subsection{Implementation Details and Evaluation Metrics}
\subsubsection{Shadow Enhancement Experiment}
\
\newline
\indent The initial values of $U,V,K,\Omega ,\Lambda ,N,{{A}_{i}},{{B}_{i}},\eta $ in the SE-BSFV algorithm should be defined. For the subspace $\{U,V\}$ of the data matrix $X$, we first use the principal component analysis (PCA) \cite{r28} method to calculate the initial subspace $\{U,V\}$, and the initial ${{A}_{i}}$, ${{B}_{i}}$ can be calculated. According to the research in references \cite{r23} and \cite{8048348}, we set $K=5$ and the nonnegative regularization parameter $\eta = 0.98$ in the following experiments to ensure the academic rigor and reliability of the experimental results. When $K=5$, the mean set of initial Gaussian components is $\{{{N}_{1}},\cdots ,{{N}_{5}}\}=\{0,\cdots ,0\}$, the initial weight set is $\{{{\pi }_{1}},\cdots ,{{\pi }_{5}}\}=\{\frac{1}{5},\cdots ,\frac{1}{5}\}$, then the initial foreground matrix $X-U{{V}^{T}}$ can be calculated according to the initial $U$ and $V$, and the initial variance set $\{\sigma _{1}^{2},\cdots ,\sigma _{5}^{2}\}$ of Gaussian components is calculated by \ref{eq16}.

In the evaluation metrics, we select Entropy and Contrast to assess the performance of shadow enhancement and background suppression.

{\bf Entropy} reflects the probability distribution of pixel gray values within an image. A higher Entropy indicates a richer information content in the image, while a lower Entropy signifies reduced information. For the shadow enhancement experiment, a lower Entropy means better performance of the algorithm as it effectively suppresses the background of the ViSAR image. The calculation for Entropy is as follows \cite{r29}:
\begin{equation}
\tag*{(25)}
\label{eq25}
Entropy=\sum\limits_{i=0}^{255}{{{P}_{i}}{{\log }_{2}}{{P}_{i}}}
\end{equation}
where ${{P}_{i}}=\frac{{{n}_{i}}}{M\times N}$ represents the probability of each gray level $i$, $M$ and $N$ denote the length and width of the image, and ${{n}_{i}}$ denotes the number of pixels with gray level $i$ in the image.

{\bf Contrast} represents the degree of variation from the lowest to the highest gray level in the image. A higher Contrast indicates a more effective shadow enhancement and background suppression by the algorithm. In this study, we used the standard deviation of the image histogram to quantify Contrast, where a 60*60 region centered on the moving target’ shadow is used for calculation. The calculation formula is as follows \cite{r30}:
\begin{equation}
\tag*{(26)}
\label{eq26}
Contrast=\sqrt{\frac{\sum\nolimits_{i=1}^{L}{{{({{n}_{i}}-\mu )}^{2}}}}{L}}
\end{equation}
where $L$ is the number of gray levels, and $\mu =\frac{1}{L}\sum\nolimits_{i=1}^{L}{{{n}_{i}}}$. 

\subsubsection{Shadow Detection Experiments}
\
\newline
\indent In the traditional detection algorithm experiment, we use the threshold segmentation algorithm based on Tsallis entropy from the reference \cite{r31} and the the spot detector algorithm from the reference \cite{985719} to detect moving targets' shadows. In addition, we add connected region processing to the flow of the two traditional detection algorithms to generate detection boxes, and delete the shadow regions whose connected regions are greater than 300 pixels and less than 50 pixels. When the IoU between the detection box and the ground truth is greater than 0.5, the target is considered to be correctly detected. Furthermore, we select 100 ViSAR images with 563 moving targets' shadows for validation.

For the deep learning-based detection experiment, we use three representative detection networks: Faster R-CNN (backbone: ResNet50 with feature pyramid network) \cite{r32}, Yolov8-m \cite{r33}, and Detection Transformer (DETR, backbone: ResNet50) \cite{r34}. These networks have been pre-trained on the COCO dataset. In this experiment, the IoU threshold is set to 0.5, and the confidence threshold is set to 0.5.

According to previous experience, we train the Yolov8-m and Faster R-CNN for 50 epochs using stochastic gradient descent (SGD) \cite{r35}. The momentum is 0.937, the weight decay is $5{{\text{e}}^{-4}}$, the initial learning rate is 0.01, and the batch size is 8. The NMS threshold for Yolov8-m is 0.7, and that for Faster R-CNN is 0.5. We also train DETR for 100 epochs using Adam \cite{r36}, the weight decay is 0.0001, the initial learning rate is $1{{e}^{-5}}$, and the batch size is 8. Furthermore, we divide the dataset according to the ratio of 8:2 for training and testing. That is, there are 720 ViSAR images in the training set and 180 ViSAR images in the test set.

In the evaluation metrics for the detection experiments, we select Precision, Recall, AP, and F1 score to assess detection performance. Ultimately, we generate the corresponding PR curves for a comprehensive comparison of the detection performance across different pre-processing algorithms.

{\bf Precision} represents the proportion of the number of correct detections to the total number of detected targets, which reflects the accuracy of detection. Precision is calculated as follows:
\begin{equation}
\tag*{(27)}
\label{eq27}
Precision=\frac{{{N}_{tp}}}{{{N}_{tp}}+{{N}_{fp}}}
\end{equation}
where ${{N}_{tp}}$ is the total number of correct detections, and ${{N}_{fp}}$ is the total number of false alarms.

{\bf Recall} represents the proportion of the number of correct detections to the number of labeled targets, and it reflects how many true targets are detected, i.e., the detection rate. It is calculated as follows:
\begin{equation}
\tag*{(28)}
\label{eq28}
Recall=\frac{{{N}_{tp}}}{{{N}_{g}}}
\end{equation}
where ${{N}_{g}}$ is the number of labeled targets in the test set.

{\bf AP} is the average of the Precision over different Recall rates.

{\bf F1 score} is commonly used as a metric for evaluating the overall performance of detection. Since Precision and Recall are usually not optimal at the same time, to comprehensively evaluate the detection performance, we introduced the F1 score for objective evaluation. F1 score is calculated as follows:
\begin{equation}
\tag*{(29)}
\label{eq29}
F1=\frac{2\times Recall\times Precision}{Recall+Precision}
\end{equation}

\subsection{Experimental Results and Analysis}
\subsubsection{Experiment on Shadow Enhancement and Background Suppression}
\
\newline
\indent
The SE-BSFV algorithm is an online learning algorithm. Although the final shadow enhancement results are output frame by frame, the subspace learning with different numbers of images will affect the performance of the algorithm. Table \ref{tab1} compares the shadow enhancement performance for different numbers of images. As shown in Table \ref{tab1}, when the number of images is set to 100, both Contrast and Entropy reach their optimal values while minimizing processing time. Therefore, in subsequent experiments, the SE-BSFV algorithm will be used to process ViSAR data every 100 images.
\begin{table}[h]
\begin{center}
\caption{Shadow Enhancement Performance for Different Numbers of Images}
\label{tab1}
\begin{tabular}{ c c c c }
\toprule
Number & Entropy$\downarrow$ & Contrast$\uparrow$ & Time(s)$\downarrow$\\
\midrule
50 & 1.1147 & 167.1366 & 2340 \\
100	& {\bf0.9285}	& {\bf184.5844}	& {\bf2322}\\
150	& 1.0258	& 169.6814	& 2473\\
\bottomrule
\end{tabular}
\end{center}
\end{table}

\begin{figure*}
\centering
\includegraphics[width=1\textwidth]{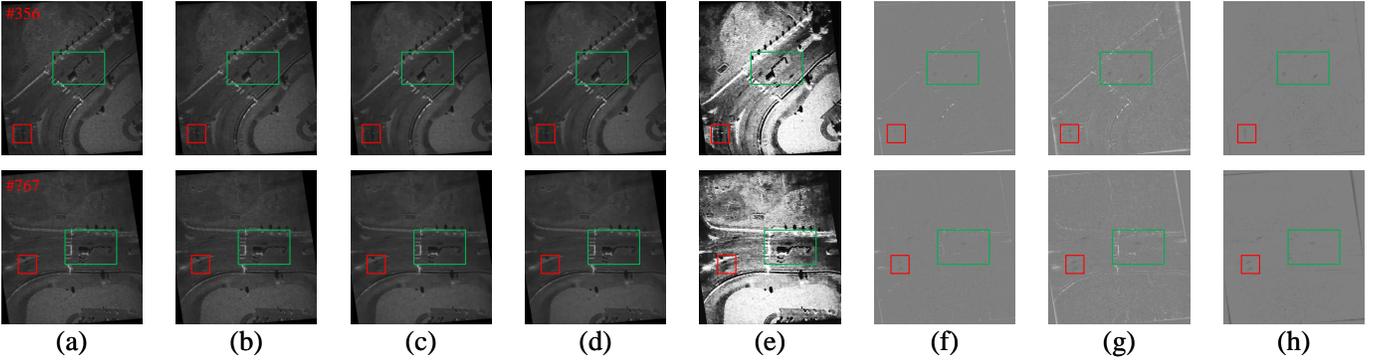}
\caption{Visualization results processed by different pre-processing algorithms. (a) Original images. (b) The MF algorithm. (c) The SAR-BM3D algorithm. (d) The V-BM3D algorithm. (e) The HESE algorithm. (f) The LRSD algorithm. (g) The SBN-3D-SD algorithm. (h) The SE-BSFV algorithm.}
\label{figure2}
\end{figure*}

\begin{figure*}
\centering
\includegraphics[width=1\textwidth]{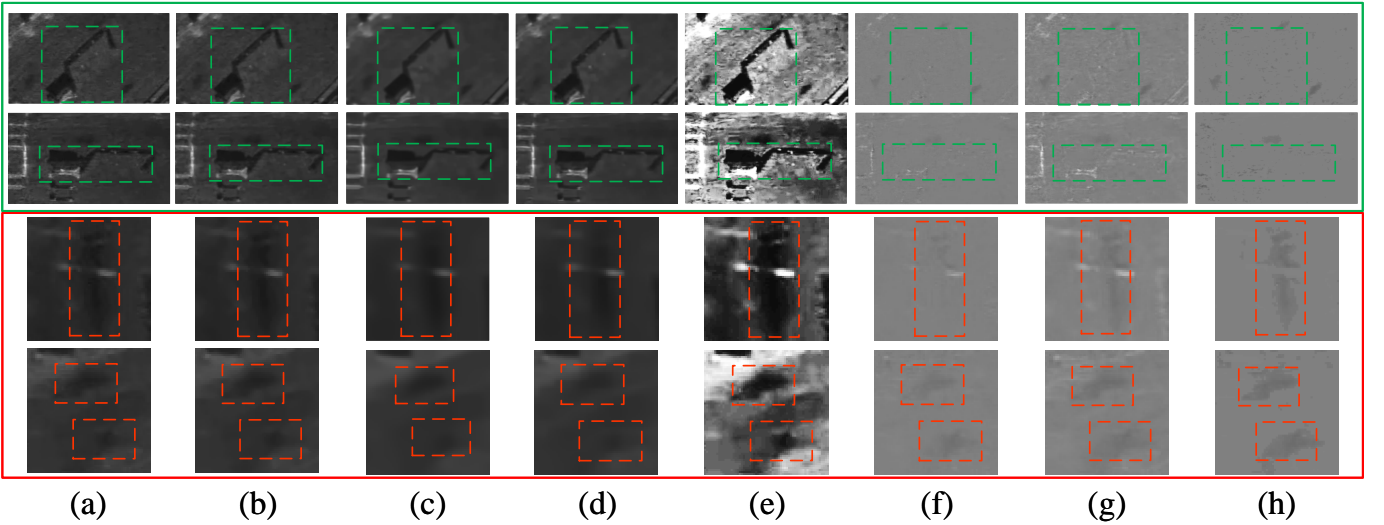}
\caption{Enlarged details of red and green boxes from different pre-processing algorithms in Fig.2. (a) Original images. (b) The MF algorithm. (c) The SAR-BM3D algorithm. (d) The V-BM3D algorithm. (e) The HESE algorithm. (f) The LRSD algorithm. (g) The SBN-3D-SD algorithm. (h) The SE-BSFV algorithm.}
\label{figure3}
\end{figure*}

Table \ref{tab2} shows the quantitative results of the different pre-processing algorithms, and Fig. \ref{figure2} presents the visualization results processed by these different pre-processing algorithms. Additionally, Fig. \ref{figure3} shows the enlarged details of the red and green boxes from the different methods in Fig.\ref{figure2}. In Fig. \ref{figure3}, the green dashed boxes are the location of the background interference, and the red dashed boxes are the location of the moving targets' shadows.

\begin{table}[h]
\begin{center}
\caption{Quantitative Results for Different Pre-Processing Algorithms (Note: "Dvf" Stands for "Designed for Visar")}
\label{tab2}
\begin{tabular}{ c c c c }
\toprule
Algorithm & DFV & Entropy$\downarrow$ & Contrast$\uparrow$\\
\midrule
Original & - & 5.6542 & 56.108 \\
MF \cite{r14} & × & 5.6321 & 57.5117\\
SAR-BM3D \cite{r15} & $\checkmark$	& 5.6324	& 67.7319\\
V-BM3D \cite{r16} &×&	5.6475&	61.653\\
HESE \cite{r13} &	×	&5.3499&	56.3612\\
LRSD \cite{r17} & $\checkmark$ & 2.3808&	151.5698\\
SBN-3D-SD \cite{r18}& $\checkmark$	&3.7119&	69.7508\\
SE-BSFV	& $\checkmark$&	{\bf0.9285}&	{\bf184.5844}\\
\bottomrule
\end{tabular}
\end{center}
\end{table}

From Fig. \ref{figure2}(a) and Fig. \ref{figure3}(a), it is evident that the original image contains a lot of background information, making it difficult to distinguish the moving targets' shadows from the complex background. At the same time, it can be seen from Table \ref{tab2} that the Entropy of the original images is the highest, reaching 5.6542, and the Contrast is the lowest, reaching 56.108. This indicates that it is difficult for human vision or MTD task to directly distinguish background and moving targets' shadows from the original image.

Fig. \ref{figure2}(b), (c), and (d), as well as Fig. \ref{figure3}(b), (c), and (d) show the results processed by the MF, SAR-BM3D, and V-BM3D algorithms, respectively. As depicted in the detailed figure of Fig. \ref{figure3}, it is evident that while these algorithms effectively reduce speckle noise in ViSAR images to a certain degree, they do not eliminate static objects, static objects' shadows, and strong scattering objects. Furthermore, there is no enhancement of moving targets' shadows. Concurrently, based on the quantitative results presented in Table \ref{tab2}, it can be observed that the Entropy processed by these algorithms are almost unchanged compared to the original images, while the Contrast only shows a notable improvement after processing with the SAR-BM3D algorithm—reaching 67.7319. Overall, the MF, SAR-BM3D, and V-BM3D algorithms primarily serve to mitigate speckle noise in ViSAR images but have minimal impact on shadow enhancement and background suppression.

Fig. \ref{figure2}(e) and Fig. \ref{figure3}(e) illustrate the results processed by the HESE algorithm. The HESE algorithm primarily approximates the histogram distribution of the original ViSAR image to a uniform distribution, thereby enhancing overall image contrast. However, it overlooks the low-rank characteristics of the background and the sparsity of moving targets' shadows. Consequently, as evidenced in Fig. \ref{figure3}(e), the HESE algorithm processing amplifies both the static objects' shadows and noise, resulting in the inadequate distinction between moving targets' shadows and background. While enhancing moving targets' shadows, the HESE algorithm inadvertently diminishes the Contrast in relevant areas and does not effectively suppress the background. As shown in Table \ref{tab2}, after the HESE processing, the Entropy and Contrast are 5.3499 and 56.3612, respectively. The improvement over the original images is very small. In summary, it can be concluded that the HESE algorithm is not a good choice for shadow enhancement and background suppression in ViSAR data.

Fig.  \ref{figure2}(f) and Fig.  \ref{figure3}(f) show the results processed by the LRSD algorithm, as well as Fig.  \ref{figure2}(g) and Fig.  \ref{figure3}(g) present the results processed by the SBN-3D-SD algorithm. Compared to the original images and the results of algorithms (b) to (e), the LRSD and SBN-3D-SD algorithms effectively mitigate background interference and reduce the stationary objects' shadows, while also preserving edge features associated with moving targets' shadows to a certain degree. The quantitative results in Table \ref{tab2} indicate that the performances of the LRSD and SBN-3D-SD algorithms are better than that of the previous four algorithms, and the Contrast after the LRSD processing is improved by 95.4618 compared with the original images, and the Contrast after the SBN-3D-SD processing is improved by 13.6428 compared with the original images. This further proves that the LRSD and SBN-3D-SD algorithms can enhance the moving targets' shadows to some extent. In addition, the Entropy after the LRSD and SBN-3D-SD algorithms processing are lower than those of the original images, which further proves that the LRSD and SBN-3D-SD algorithms suppress the background of ViSAR images. From the comparison of the detailed images in Fig.  \ref{figure3}(f) and (g), the background of the LRSD processed image is suppressed, and the moving targets' shadows are indeed enhanced to some extent compared with the original image. But the edge features, area, and shape of moving targets' shadows are not as well-preserved as in the images after the SBN-3D-SD processing.

Fig.  \ref{figure2}(h) presents the results after the SE-BSFV processing, and Fig.  \ref{figure3}(h) illustrates the corresponding detailed images. As evidenced in Fig.  \ref{figure2}(h) and Fig.  \ref{figure3}(h), the distinction between the moving targets' shadows and the background is markedly enhanced, with a more pronounced suppression of both the background and stationary objects' shadows compared to the LRSD and SBN-3D-SD algorithms. Furthermore, as depicted in Fig.  \ref{figure3}(h), the edge features, area, and shape of moving targets' shadows are also much easier to discern than those observed in original images and any of the above six pre-processing algorithms. The quantitative results presented in Table \ref{tab2} indicate that the Contrast after SE-BSFV processing is substantially improved—128.4764 higher than that of the original images, 128.2232 higher than the HESE processing, 33.0146 higher than the LRSD processing, and an increase of 114.8336 over the SBN-3D-SD processing. This indicates that it is easier to distinguish background and moving targets' shadows after the SE-BSFV processing. Additionally, the Entropy is the lowest among all algorithms at 0.9285, further substantiating that the SE-BSFV algorithm exhibits superior performance in background suppression.

In conclusion, as illustrated in Fig.  \ref{figure2}, Fig.  \ref{figure3}, and Table \ref{tab2}, the SE-BSFV algorithm significantly enhances the distinction between the moving targets' shadows and the background compared to other pre-processing algorithms. This could have a positive impact on the performance of MTD.

\begin{figure}[h]
\centering
\includegraphics[width=\columnwidth]{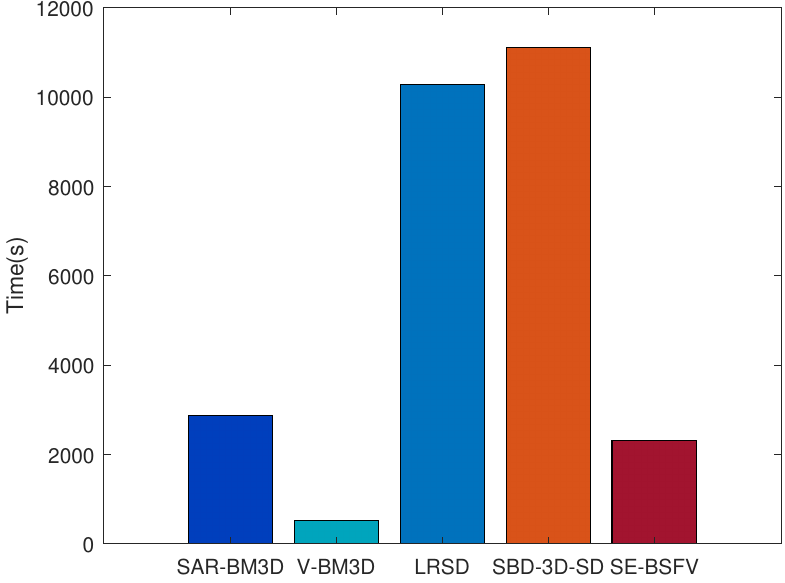}
\caption{The processing time of the different pre-processing algorithms on the SNL's ViSAR data.}
\label{fig4}
\end{figure}

Fig. \ref{fig4} shows the processing time of the five pre-processing algorithms applied to the SNL's ViSAR data. The running time of the SE-BSFV algorithm is only slightly longer than that of the V-BM3D algorithm, 558s shorter than that of the SAR-VBM3D algorithm, 7956s shorter than that of the LRSD algorithm, and 8802s shorter than that of the SBN-3D-SD algorithm. Therefore, the SE-BSFV algorithm can achieve better shadow enhancement and background suppression performance with less processing time and does not affect the efficiency of subsequent detection tasks.

\begin{table*}[htbp]
\begin{center}
\caption{Quantitative Results Under Different Noise Sizes}
\label{tab3}
\begin{tabular*}{0.72\linewidth}{c c c | c c c | c c c}
\toprule
Algorithm & SNR & Entropy$\downarrow$ & Algorithm & SNR & Entropy$\downarrow$ & Algorithm & SNR & Entropy$\downarrow$\\
\midrule
Original &	\multirow{8}{*}{5dB}	&6.8305	&Original&	\multirow{8}{*}{10dB}	&6.0147&	Original &	\multirow{8}{*}{15dB}	&5.8291\\
     MF	&	  &6.1382	&MF	& &	5.7898&	MF&	&	5.7214\\
SAR-BM3D	& &	5.6803	&SAR-BM3D	& &	5.6374	&SAR-BM3D &	&	5.6302\\
V-BM3D	& &	5.6738	&V-BM3D	& &	5.6449	&V-BM3D &	&	5.6478\\
HESE&	&	7.5636&	HESE	& &	7.9057&	HESE&	 &	7.9347\\
LRSD&	&	6.7188&	LRSD&	&	5.0151&	LRSD&	&	4.1619\\
SBN-3D-SD&	&	6.7304	&SBN-3D-SD&	&	5.1764&	SBN-3D-SD&	&	4.5198\\
SE-BSFV&	&	{\bf 3.5686}	&SE-BSFV&	&	{\bf 3.3179}	&SE-BSFV&	&	{\bf 3.2008}\\
\bottomrule
\end{tabular*}
\end{center}
\end{table*}

To further verify the robustness of the SE-BSFV algorithm, the white Gaussian noise is introduced into the original ViSAR images for analysis at SNR of 5 dB, 10 dB, and 15 dB, respectively. Since the moving target's shadow is an area with a low gray value, introducing the white Gaussian noise with a gray value much larger than the moving target's shadow will result in the inability to accurately calculate the Contrast. In this experiment, the Contrast is not used for evaluation. The quantitative results under different noise sizes are presented in Table \ref{tab3}. As indicated in Table \ref{tab3}, as the SNR decreases, the Entropy of the images will increase. Regardless of the SNR, the Entropy after the SE-BSFV processing is the smallest. This further substantiates that the SE-BSFV algorithm exhibits greater robustness.

\subsubsection{Experiments on Shadow Detection}
\
\newline
\indent To further evaluate the impact of different pre-processing algorithms on MDT, we assessed their performance using traditional and deep learning-based detection algorithms.

\textit{a) Detection Experiment with Traditional Algorithms:}

Table \ref{tab4} shows the quantitative results based on the algorithms of references \cite{r31} and \cite{985719}. As illustrated in Table \ref{tab4}, in the quantitative results of the two algorithms, the original images and the images processed by the MF, SAR-BM3D, V-BM3D, and HESE algorithms have lower Precision and Recall, while the images processed by the LRSD, SBN-3D-SD, and SE-BSFV algorithms have higher Precision and Recall. This indicates that the LRSD, SBN-3D-SD, and SE-BSFV algorithms with background suppression function contribute positively to the performance of traditional detection algorithm. Furthermore, the SE-BSFV algorithm has the highest Recall rate, reaching 74.2\% and 49.4\% respectively. Although its Precision is not the highest, its F1 score reaches 0.548 and 0.208 respectively—surpassing that of all other algorithms.  Therefore, this proves that the SE-BSFV algorithm effectively improves the performance of the traditional detection algorithms more than other pre-processing algorithms in this study.

\begin{table*}[ht]
\begin{center}
\caption{Quantitative Results of Detection for Traditional Algorithms}
\label{tab4}
\begin{tabular*}{0.75\linewidth}{c c c c c c c c c}
\toprule
\multicolumn {2}{c}{Method} & ${{N}_{g}}$ & ${{N}_{tp}}$  $\uparrow$ & ${{N}_{fp}}$ $\downarrow$ & ${{N}_{fn}}$ $\downarrow$ & Precision (\%) $\uparrow$ & Recall (\%) $\uparrow$ & F1 $\uparrow$\\
\midrule
\multirow {8}{*}{Reference \cite{r31}} & Original &	\multirow{8}{*}{563}	&139 &	4265	& 424	&3.2	&24.7	&0.057\\
&MF &		&142&	4171&	421	&3.3&	25.2	&0.058\\
&SAR-BM3D	&	&140&	2902&	423&	4.6&	24.9	&0.078\\
&V-BM3D&		&124	&2956&	439&	4.0&	22.0&	0.068\\
&HESE&		&16&	3463&	547	&0.46&	2.8&	0.008\\
&LRSD&		&323&	422&	240&	43.4&	57.4&	0.494\\
&SBN-3D-SD	&	&228&	\textbf{239}&	335&	\textbf{48.8}&	40.5&	0.443\\
&SE-BSFV&		&{\bf418}&	542&	{\bf145}&	43.5&	{\bf74.2}&	{\bf0.548}\\
\bottomrule
\multirow {8}{*}{Reference \cite{985719}} & Original &	\multirow{8}{*}{563}	& 89 &	4987	& 474	& 1.8	&15.8	&0.032\\
&MF &		&93&	5026 & 470	&1.8 &	16.5	&0.032\\
&SAR-BM3D	&	&99&	2793&	464 &	3.4 &	17.6	&0.057\\
&V-BM3D&		&102	&2714&	461&	3.6  &	18.1 &	0.060\\
&HESE&		&15&	3985 & 548	&0.38&	2.7&	0.007\\
&LRSD&		&68&	\textbf{163}&	495&	\textbf{29.4}&	12.1&	0.171\\
&SBN-3D-SD&	&233&	1841&	330&	11.2 &	41.4&	0.176\\
&SE-BSFV&		&{\bf278}&	1827&	{\bf285}&	13.2 &	{\bf49.4}&	{\bf0.208}\\
\bottomrule
\end{tabular*}
\end{center}
\end{table*}

\begin{table*}[ht]
\begin{center}
\caption{Quantitative Results of Detection for Deep Learning-Based Detectors}
\label{tab5}
\begin{tabular*}{0.83\linewidth}{c c c c c c c c c c}
\toprule
\multicolumn {2}{c}{Method} & ${{N}_{g}}$ & ${{N}_{tp}}$  $\uparrow$ & ${{N}_{fp}}$ $\downarrow$ & ${{N}_{fn}}$ $\downarrow$ & Precision (\%) $\uparrow$ & Recall (\%) $\uparrow$ & AP@0.5 $\uparrow$ & F1 $\uparrow$\\
\midrule
\multirow {8}{*}{Yolov8-m}  & Original &\multirow {8}{*}{912}	&447&	111&	465&	80.1&	49&	0.58	&0.61\\
 &  MF& 		&452&	104&	460	&81.3&	49.6&	0.59	&0.62\\
 &  SAR-BM3D&		&451&	60&	461&	88.3&	49.5&	0.63&	0.63\\
 &  V-BM3D&		&302&	89&	610&	77.2&	33.1&	0.39&	0.46\\
 &  HESE	&	&468&	111&	444&	80.8&	51.3&	0.64&	0.63\\
 &  LRSD&		&587&	58	&325	&91&	64.4	&0.73	&0.75\\
 &  SBN-3D-SD&		&636&	{\bf36} &	276&	{\bf94.6}&	69.7&	0.83&	0.8\\
 &  SE-BSFV&		&{\bf684}&	62&	{\bf228} &	91.7&	{\bf75}&	{\bf0.84}&	{\bf0.83}\\
\bottomrule
\multirow {8}{*}{Faster R-CNN}  & Original &\multirow {8}{*}{912}	&745	&378	&167	&66.3&	81.7&	0.71	&0.73\\
&MF &		&719&	188&	194&	79.3&	78.8	&0.73&	0.79\\
&SAR-BM3D&		&781&	323&	131&	70.7	&85.6&	0.71&	0.77\\
&V-BM3D&		&585&	211&	327&	73.5&	64.1&	0.48	&0.68\\
&HESE	&	&738&	306&	174	&70.7&	80.9&	0.72&	0.75\\
&LRSD	&	&757&	185&	155&	80.4&	83&	0.77&	0.82\\
&SBN-3D-SD&		&814&	{\bf133}&	98&	{\bf86}&	89.3&	0.87&	{\bf0.88}\\
&SE-BSFV&		&{\bf837}&	200&	{\bf75}&	80.7&	{\bf91.8}&	{\bf0.90}&	0.86\\
\bottomrule
\multirow {8}{*}{DETR}  & Original &\multirow {8}{*}{912}	&799&	404	&113&	66.4	&87.6&	0.65	&0.76\\
&MF &		&799&	235&	113&	77.3&	87.6	&0.71&	0.82\\
&SAR-BM3D&	&	802	&263&	110&	75.3&	87.9	&0.69&	0.81\\
&V-BM3D&		&734&	352	&178	&67.6	&80.5	&0.52	&0.73\\
&HESE&		&758&	307&	154&	71.2&	83.1	&0.68&	0.77\\
&LRSD	&	&789&	288&	123&	73.3&	86.5&	0.74&	0.79\\
&SBN-3D-SD&		&{\bf866}	&312	&{\bf45}&	73.5&	{\bf95}&	0.82&	{\bf0.83}\\
&SE-BSFV&		&856&	{\bf297}&	56&	{\bf74.2}	&93.9	&{\bf0.89}	&{\bf0.83}\\
\bottomrule
\end{tabular*}
\end{center}
\end{table*}

\textit{b) Detection Experiment with Deep Learning-based Detectors:}

\begin{figure*}[ht]
\centering
\includegraphics[width=1\textwidth]{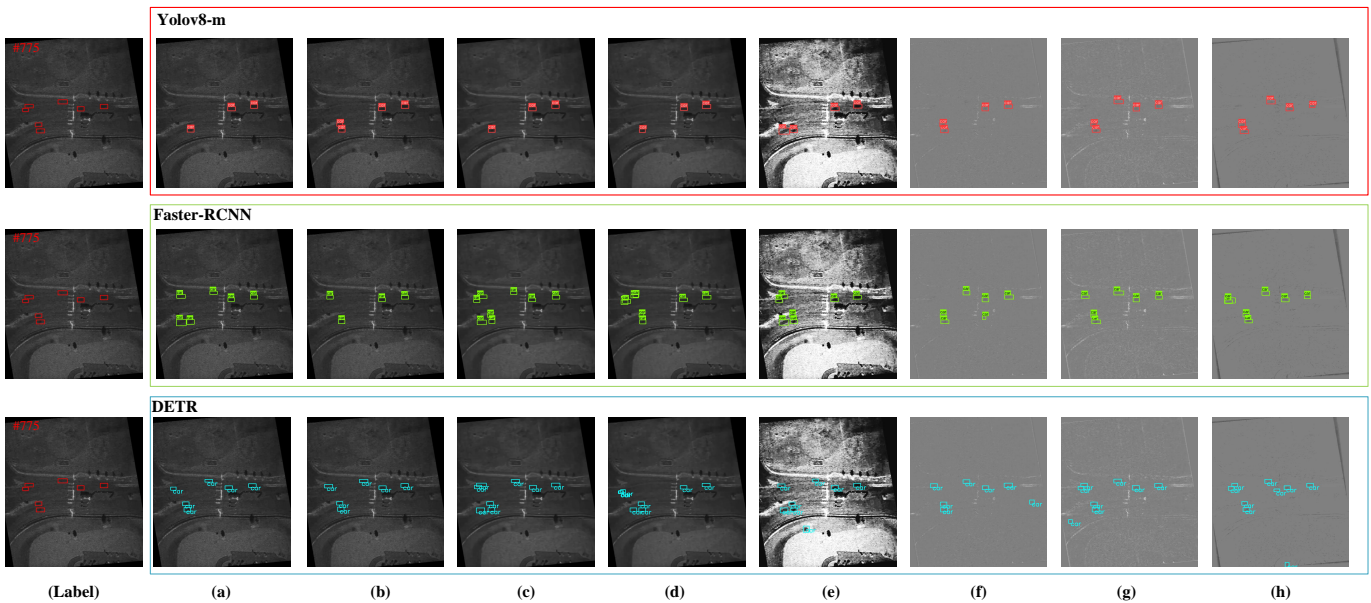}
\caption{The visualization results. (a) Original image detection results. (b) Detection results after the MF processing. (c) Detection results after the SAR-BM3D processing. (d) Detection results after the V-BM3D processing. (e) Detection results after the HESE processing. (f) Detection results after the LRSD processing. (g) Detection results after the SBN-3D-SD processing. (h) Detection results after the SE-BSFV processing.}
\label{figure5}
\end{figure*}

To further evaluate the impact of different pre-processing algorithms on the performance of deep learning-based detectors, we conducted a detection experiment using Yolov8-m, Faster R-CNN, and DETR. Table \ref{tab5} shows the quantitative results of detection for deep learning-based detectors, and Fig. \ref{figure5} presents the visualization results of the Yolov8-m, Faster R-CNN, and DETR in frame 775 on the SNL's ViSAR data.

The quantitative results of the Yolov8-m detector in Table \ref{tab5} indicate a significant number of missing and false alarms when directly detecting the original images. This issue arises because the gray values of the background surrounding moving targets' shadows closely resemble those of the shadows themselves, which causes the detection network to fail to fully extract and characterize the moving targets' shadows. This result is further corroborated by the visualization results of the Yolov8-m shown in Fig. \ref{figure5}. After the MF, SAR-BM3D, and HESE processing, the missing and false alarms are suppressed in the Yolov8-m detector, and the F1 score is also improved compared with the original images. However, it is noteworthy that the detection results processed by the V-BM3D algorithm are worse than those of the original images. This may be due to a lack of targeted optimization for ViSAR data within the V-BM3D algorithm. After the LRSD processing, there are fewer missing and false alarms than the above four algorithms, the Precision reaches 91\%, and the Recall rate reaches 64.4\%. After the SBN-3D-SD algorithm processing, the Precision is highest among all algorithms, reaching 94.6\%, but the recall rate is only 69.7\%. In contrast, after the SE-BSFV algorithm processing, the Recall rate is the highest among the other algorithms, reaching 75\%. Even though the Precision is a little lower than that of the SBN-3D-SD algorithm, the F1 score reaches 0.83, which is 0.03 higher than that of the SBN-3D-SD algorithm and 0.22 higher than that of the original images, which is the best among these algorithms.

\begin{figure*}[ht]
\centering
\includegraphics[width=1\textwidth]{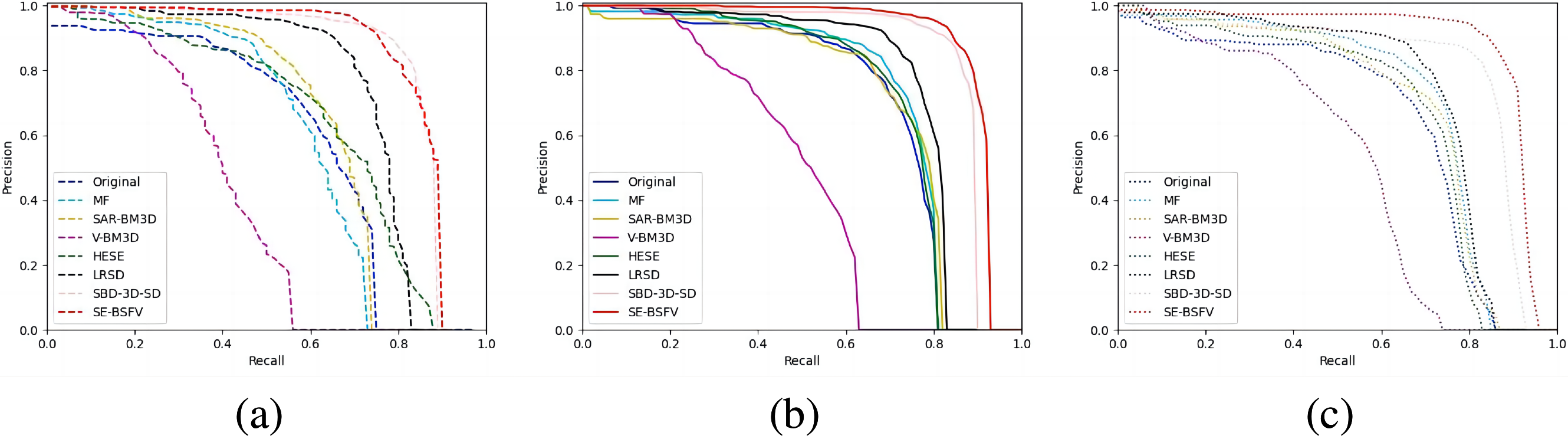}
\caption{The PR curves of different pre-processing algorithms on different detectors. (a) Yolov8-m. (b) Faster R-CNN. (c) DETR.}
\label{figure6}
\end{figure*}
Unlike the Yolov8-m detector, which is a one-stage detector, the Faster R-CNN detector is a two-stage detector that needs to generate proposals and then identify the real targets from these proposals. Consequently, the Recall of the Faster R-CNN detector tends to be higher than that of the Yolov8-m detector. As shown in Table \ref{tab5} and illustrated in Fig. \ref{figure5}, the Recall of the Faster R-CNN is improved over that of Yolov8-m, but the higher Recall is obtained at the cost of decreased Precision. Furthermore, the trend of detection results of different pre-processing algorithms in the Faster-RCNN detector is similar to the trend of the Yolov8-m detector. Notably, after the SE-BSFV processing, the Recall reaches the highest value among all the comparison algorithms, which is 91.8\%, an increase of 10.1 \% compared with the original images. While the Precision is not the best, being 5.3\% lower than that of the SBN-3D-SD algorithm, it still remains superior to the other five algorithms. Moreover, the F1 score of the SE-BSFV algorithm is only 0.02 lower than that of the SBN-3D-SD algorithm.

The DETR detector is a relatively new target detection algorithm that combines CNNs with a Transformer to directly predict the final detection results. In the visualization results of Fig. \ref{figure5}, all targets are correctly detected in Fig. \ref{figure5}(h), although some false alarms are present. From the quantitative detection results for DETR shown in Table \ref{tab5}, the trend of detection results of different pre-processing algorithms is also similar to the Yolov8-m and Faster R-CNN detectors. The slight difference is that after the SE-BSFV algorithm processing, the Precision is the highest among all pre-processing algorithms, reaching 74.2\%. The Recall rate is 93.4\%, which is 1.1\% lower than the SBN-3D-SD algorithm, and the F1 score is 0.83, which is comparable to the SBN-3D-SD algorithm.

Fig. \ref{figure6} shows the PR curves of different pre-processing algorithms on different detectors at IoU = 0.5. The SE-BSFV algorithm has better PR curves than the other pre-processing algorithms on all detectors, and these results are also reflected in the AP values shown in Table \ref{tab5}. This further confirms that the detection performance after the SE-BSFV processing is significantly improved.

In general, the detection experiment with deep learning-based detectors demonstrates that the detection performance of all detectors is significantly improved after the SE-BSFV algorithm processing, indicating that the SE-BSFV algorithm has a certain versatility for various object detection networks. In addition, on the Yolov8-m and DETR detectors, the SE-BSFV algorithm has a greater performance improvement than the SBN-3D-SD algorithm, and the SE-BSFV algorithm runs 5 times faster than the SBN-3D-SD algorithm. Consequently, the SE-BSFV algorithm can substantially improve the detection performance of deep learning-based detectors without compromising detection speed.

\section{Conclusion}
In this study, a novel pre-processing algorithm called SE-BSFV is proposed to enhance the moving targets' shadows and suppress the background in ViSAR images, thereby improving the performance of MTD. Our analysis reveals that the ViSAR data can be decomposed into a low-rank matrix representing the background and a sparse matrix corresponding to the moving targets' shadows. Additionally, the ViSAR data can be effectively modeled using GMD. To leverage these characteristics, the SE-BSFV algorithm first employs the SURF algorithm to register the ViSAR images and then models these images with GMD. Subsequently, the knowledge acquired from the previous frames is utilized to estimate GMD parameters for the current frame. The EM algorithm is then applied to determine model subspace parameters, updating the current frame's low-rank matrix while extracting its corresponding sparse matrix. Finally, the remaining strong scattering objects in the sparse matrix are eliminated by the ADMM processing, and the final smooth images containing the moving targets’ shadows are obtained. The shadow enhancement experimental results demonstrate that the SE-BSFV algorithm can effectively enhance the moving targets’ shadows and suppress the background in ViSAR data. Furthermore, compared with other pre-processing algorithms, it exhibits greater robustness and speed. The detection experimental results reveal that, whether in the traditional detection algorithm or the deep learning-based detectors, compared with other pre-processing algorithms, the SE-BSFV algorithm can more effectively improve the detection performance while ensuring efficiency.

In future studies, we will further explore the essential differences between the moving targets' shadows and the noise that still exists on the edges of roads or at the edges of strong scattering objects, in order to extract purer moving target shadows.

\section{Appendix}
\subsection{The Derivation of the EM Algorithm in the GMD-LRR Algorithm}
\label{Appendix-A}
The EM algorithm in the GMD-LRR algorithm is derived as follows:

1) E-step:
    
The hidden variable associated with the ${{x}_{ij}}$ is defined as $z$, and the hidden distribution $q({{z}_{k}})={{\gamma }_{ijk}}$ represents the probability that each data in the GMD comes from the $k\in \left\{ 1,\ldots ,K \right\}$ distribution. Then the hidden variable has a discrete value set $Z=\left\{ {{z}_{1}},\ldots ,{{z}_{K}} \right\}$, and the hidden distribution ${{\gamma }_{ijk}}$ is calculated in the E-step, that is:

\begin{equation}
\tag*{(A.1)}
\label{eqA.1}
q({{z}_{k}})={{\gamma }_{ijk}}=\frac{{{\pi }_{k}}P\left[ {{x}_{ij}}\left| {{u}_{i}}{{({{v}_{i}})}^{T}},\sigma _{k}^{2} \right. \right]}{\sum\nolimits_{k=1}^{K}{{{\pi }_{k}}P\left[ {{x}_{ij}}\left| {{u}_{i}}{{({{v}_{i}})}^{T}},\sigma _{k}^{2} \right. \right]}} 
\end{equation}

2) M-step:

The M-step solves the maximization upper bound of the parameters $U,V,\Omega ,\Lambda $ in the E-step:

\begin{strip}
\begin{equation} \tag*{(A.2)}
\label{eqA.2}
\begin{split}
\ln p(X,Z\left| U,V,\Omega ,\Lambda  \right.)&=\sum\nolimits_{i=1}^{d}{\sum\nolimits_{j=1}^{n}{\sum\nolimits_{k=1}^{K}{{{\gamma }_{ijk}}\left\{ \ln {{\pi }_{k}}-\ln \sqrt{2\pi }{{\sigma }_{k}}-\frac{{{[{{x}_{ij}}-{{u}_{i}}{{({{v}_{j}})}^{T}}]}^{2}}}{2\sigma _{k}^{2}} \right\}}}} \\ 
& =\sum\nolimits_{i=1}^{d}{\sum\nolimits_{j=1}^{n}{\sum\nolimits_{k=1}^{K}{{{\gamma }_{ijk}}\left\{ \ln {{\pi }_{k}}-\ln {{\sigma }_{k}}-\frac{{{[{{x}_{ij}}-{{u}_{i}}{{({{v}_{j}})}^{T}}]}^{2}}}{2\sigma _{k}^{2}} \right\}}}}+C  
\end{split}
\end{equation}
\end{strip}

After calculating the hidden distribution ${{\gamma }_{ijk}}$ through the E step, the parameters $U,V,\Omega ,\Lambda $ are updated alternately. Update $\Omega ,\Lambda $:
\begin{equation}
\label{eqA.3}
\left\{ \begin{array}{*{35}{l}}
   {{\pi }_{k}}=\frac{{{N}_{k}}}{\sum\nolimits_{k=1}^{K}{{{N}_{k}}}}  \\
   \sigma _{k}^{2}=\frac{1}{{{N}_{k}}}\sum\nolimits_{i,j}{{{\gamma }_{ijk}}}{{[{{x}_{ij}}-{{u}_{i}}{{({{v}_{j}})}^{T}}]}^{2}}  \\
\end{array} \right. \tag*{(A.3)}
\end{equation}
where ${{N}_{k}}=\sum\nolimits_{i,j}{{{\gamma }_{ijk}}}$.

According to \ref{eqA.2}, updating $U$ and $V$ can be solved by a weighted L2-LRR problem as follows:

\begin{equation} \tag*{(A.4)}
\label{eqA.4}
\begin{split}
& \sum\nolimits_{i,j\in \Omega }{\sum\nolimits_{k=1}^{K}{{{\gamma }_{ijk}}\left\{ -\frac{{{[{{x}_{ij}}-{{u}_{i}}{{({{v}_{j}})}^{T}}]}^{2}}}{2\sigma _{k}^{2}} \right\}}} \\ 
& =-\sum\nolimits_{i,j\in \Omega }{\left( \sum\nolimits_{k=1}^{K}{\frac{{{\gamma }_{ijk}}}{2\sigma _{k}^{2}}} \right){{[{{x}_{ij}}-{{u}_{i}}{{({{v}_{j}})}^{T}}]}^{2}}} \\ 
& =-{{\left\| W\odot (X-U{{V}^{T}}) \right\|}_{{{L}_{2}}}}
\end{split}
\end{equation}
where the corresponding position of $W=\left\{ {{w}_{i,j}} \right\}$ is:

\begin{equation} \tag*{(A.5)}
\label{eqA.5}
\begin{split}
{{w}_{i,j}}=\left\{ \begin{array}{*{35}{l}}
   \sqrt{\sum\nolimits_{k=1}^{K}{\frac{{{\gamma }_{ijk}}}{2\sigma _{k}^{2}}}}, & i,j\in \Theta   \\
   0, & i,j\notin \Theta   \\
\end{array} \right.
\end{split}
\end{equation}

The ALS method \cite{r37} is used to solve \ref{eqA.4} until the result converges, whose corresponding $U$ and $V$ are the matrix solution closest to the true value \cite{r25}:

\begin{equation}
\label{eqA.6}
{{\left\| X-U{{V}^{T}} \right\|}_{{{L}_{2}}}}\to \min \sum\nolimits_{i,j\in \Omega }{\left[ {{x}_{ij}}-{{u}_{i}}{{({{v}_{j}})}^{T}} \right]} \tag*{(A.6)}
\end{equation}

\subsection{Spatial Structure Analysis of the ViSAR Data}
\label{Appendix-B}

For the SNL's ViSAR data, its ViSAR images can be expressed as $[{{x}_{1}},\cdots ,{{x}_{i}},\cdots ,{{x}_{n}}]\in {{\Re }^{{{N}_{a}}\times {{N}_{r}}}}$, where ${{N}_{a}}$ and ${{N}_{r}}$ denote the number of pixels in the azimuth and range directions. The ViSAR data can be completely represented as a matrix $X$ by vectorizing each image of the ViSAR data and placing the corresponding vectors of multiple images side-by-side in chronological order. The CDF \cite{r38} is used to analyze the spatial structure information of the matrix $X$. The calculation formula is as follows:

\begin{equation}
\label{eqB.1}
CDF(\rho )=\frac{\sum\nolimits_{j=1}^{\rho \%\times \min (d,n)}{\Gamma (j,j)}}{\sum\nolimits_{j=1}^{\min (d,n)}{\Gamma (j,j)}} \tag*{(B.1)}
\end{equation}
where $\Gamma $ represents the diagonal matrix formed by the singular values of the matrix $X$, $\Gamma (j,j)$ represents the j-th largest singular value and $\rho \%$ represents the singular values of the first $\rho $ percent.

\begin{figure}[h]
\centering
\includegraphics[width=\columnwidth]{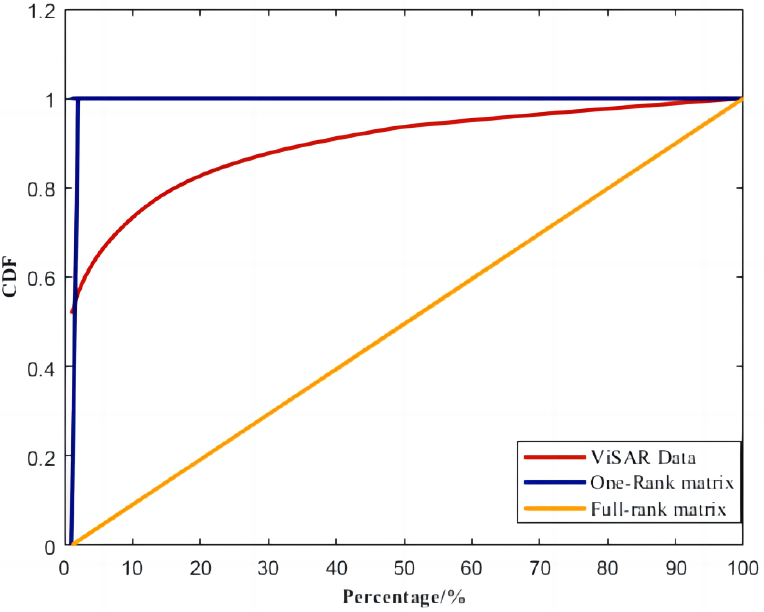}
\caption{The curve of CDF.}
\label{figure7}
\end{figure}

{\bf Low-rank property}: Fig. \ref{figure7} shows the CDF analysis, where the red line is the CDF of the matrix $X$ formed by 100 images, the blue line is the CDF cumulative distribution of a one-rank matrix, and the yellow line is the CDF cumulative distribution of a full-rank matrix. The cumulative distribution of the ViSAR data is closer to the one-rank matrix, which indicates that the information redundancy in the ViSAR data is high and the ViSAR data has a low rank.

\begin{figure}[bp]
\centering
\includegraphics[width=\columnwidth]{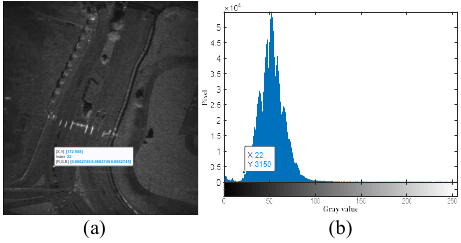}
\caption{The sparsity analysis of moving targets' shadows. (a) The pixel value at the center of the shadow. (b) Pixel statistic.}
\label{figure8}
\end{figure}

{\bf Sparsity property}: The sparsity of the matrix $X$ is analyzed by calculating the proportion of pixels of moving targets' shadows in ViSAR images \cite{r18}. Fig. \ref{figure8} shows the sparsity analysis diagram of the moving targets' shadows in a ViSAR image, Fig. \ref{figure8}(a) shows the pixel value of the center of a moving target's shadow, and Fig. \ref{figure8}(b) shows the pixel statistics of the image. As can be seen in Fig. \ref{figure8}(a), the shadow is dark and the gray value is only 22, and the distribution of moving targets' shadows is sparse. It can be seen from Fig. \ref{figure8}(b) that the moving targets' shadows occupy a small proportion in a single image. Therefore, the moving targets' shadows have sparsity in a single image.

According to the above analysis, after converting the whole ViSAR data into a two-dimensional matrix $X$, the background can be regarded as a low-rank matrix $U{{V}^{T}}$ and the moving targets' shadows can be regarded as a residual matrix (sparse matrix) $X-U{{V}^{T}}$. Therefore, we can use the concept of LRR to suppress the background and enhance the shadows to improve the distinction between moving targets' shadows and background.

\begin{figure*}[h]
\centering
\includegraphics[width=1\textwidth]{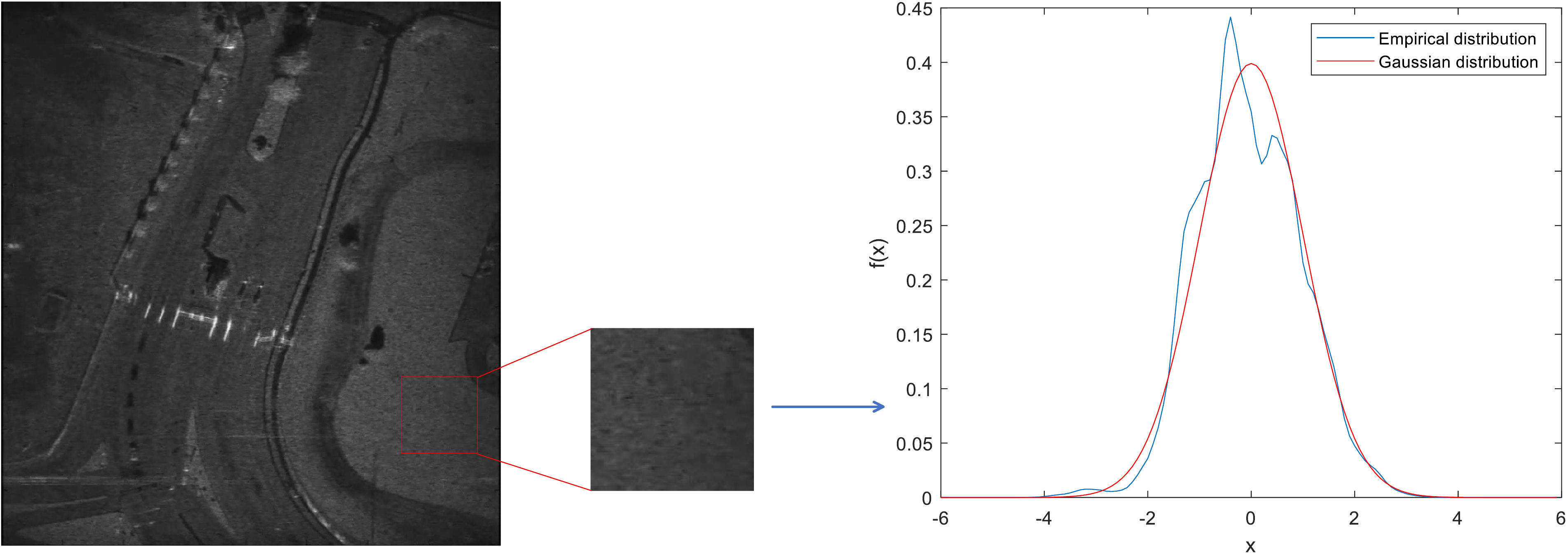}
\caption{Background distribution analysis of the ViSAR data.}
\label{figure9}
\end{figure*}

In addition, reference \cite{r39} found that during ViSAR imaging, background clutter and noise in ViSAR data can sometimes be modeled as an approximate Gaussian distribution. Fig. \ref{figure9} shows the background distribution of the ViSAR data, the red line is the standard normal distribution curve with expectation 0 and variance 1, and the blue line is the distribution of a region of a ViSAR image. The distribution of this region is in line with the Gaussian characteristics, but there is certain difference with the Gaussian distribution, and there are multiple peaks. Therefore, in the subsequent study, the GMD is planned to be adopted for LRR model modeling.

\bibliographystyle{IEEEtran}
\bibliography{IEEEabrv, references.bib}          

\end{document}